\documentclass{article}
\usepackage{amsmath,amssymb}
\usepackage{graphicx}

\newcommand{\etal}{{\it et al.}}

 \newcommand{\RR}{\mathbb{R}}

 \newcommand{\T}{{}^\top}

 \newcommand{\bm}[1]{\boldsymbol{#1}}

 \newcommand{\vz}{\boldsymbol{z}}

 \newcommand{\valpha}{\boldsymbol{\alpha}}
 \newcommand{\vbeta}{\boldsymbol{\beta}}

 \newcommand{\mK}{\boldsymbol{K}}

 \newcommand{\minimize}{\mathop{\rm minimize}}

 \newcommand{\Eqref}[1]{Eq.~{\eqref{#1}}}

 \newcommand{\Figref}[1]{Fig.~{\ref{#1}}}

\title{Sparsity-accuracy trade-off in MKL}
\author{Ryota Tomioka \& Taiji Suzuki\footnote{Both authors contributed
equality to this work.}\\\texttt{\{tomioka, t-suzuki\}@mist.i.u-tokyo.ac.jp}}
\date{}
\begin{document}
\maketitle

\begin{abstract}
\noindent We empirically investigate  the best trade-off between sparse and
 uniformly-weighted multiple kernel learning (MKL) using the elastic-net regularization on real and
 simulated datasets. We find that the best trade-off parameter depends
 not only on the sparsity of the true kernel-weight spectrum but also on the
 linear dependence among kernels and the number of samples.
\end{abstract}
\section{Introduction}
 Sparse multiple kernel learning (MKL; see~\cite{LanCriBarGhaJor04,MicPon05,BacLanJor04}) is often outperformed by the simple uniformly-weighted MKL
 in terms of accuracy~\cite{Cor09,KloBreSonLasMulZie10}. However the sparsity offered by the sparse MKL is
 helpful in understanding which feature is useful and can also save a lot
of computation in practice. In this paper we investigate this trade-off
 between the sparsity and accuracy using an elastic-net type
 regularization term which is a smooth interpolation between the
  sparse ($\ell_1$-) MKL and the uniformly-weighted MKL. In addition,
 we extend the recently proposed SpicyMKL algorithm~\cite{SuzTom09} for efficient
 optimization in  the proposed elastic-net regularized MKL
 framework. Based on real and  simulated MKL problems with more than
 1000 kernels, we show that:
\begin{enumerate}
 \item Sparse MKL indeed suffers  from poor accuracy when
       the  number of samples is small. 
 \item As the  number of samples grows larger, the  difference in the accuracy
       between sparse  MKL and uniformly-weighted  MKL becomes smaller.
 \item Often the best accuracy is obtained in between the sparse and
       uniformly-weighted MKL. 
This can be explained by the dependence among candidate kernels having
      neighboring kernel parameter values.
\end{enumerate}

\section{Method}
Let us assume that we are provided with $M$ reproducing kernel Hilbert
spaces (RKHSs) equipped with kernel functions $k_m$:
$\mathcal{X}\times\mathcal{X}\rightarrow\RR$
$(m=1,\ldots,M)$ and the task is to learn a classifier from $N$ training
examples $\{(x_i,y_i)\}_{i=1}^N$, where  $x_i\in\mathcal{X}$ and
$y_i\in\{-1,+1\}$ ($i=1,\ldots,N$).
We formulate this problem into the following minimization problem:
\begin{align}
\label{eq:functional-problem}
 \minimize_{\substack{f_m\in\mathcal{H}_m\\(m=1,\ldots,M),\\ b\in\RR}}\quad &\sum_{i=1}^N\ell\Bigl(\sum_{m=1}^mf_m(x_i)+b,y_i\Bigr)+C\sum_{m=1}^M\Bigl((1-\lambda)\|f_m\|_{\mathcal{H}_m}+\frac{\lambda}{2}\|f_m\|_{\mathcal{H}_m}^2\Bigr),
\end{align}
where in the first term, $f_m$ is a member of the $m$-th RKHS
$\mathcal{H}_m$, $b$ is a bias term, and $\ell$ is a loss function; in
this paper we use the logistic loss function.  
 The second term is a
regularization term and is a mixture of $\ell_1$- and $\ell_2$-
regularization terms. The constant $C$ ($>0$) determines the overall trade-off
between the loss term and the regularization terms.
Here the first regularization term is the linear sum of
RKHS norms, which is known to make only few $f_m$'s non-zero (i.e.,
sparse, see \cite{Tib96,YuaLin06,Bac08}); the second regularization term is the squared 
sum of RKHS norms.  The two regularization terms are balanced by the
constant $\lambda$ ($0\leq\lambda\leq 1$); $\lambda=0$ 
corresponds to sparse ($\ell_1$-) MKL and $\lambda=1$ corresponds to
uniformly-weighted MKL.

Due to the representer theorem (see~\cite{SchSmo02}), the solution of
the above minimization problem \eqref{eq:functional-problem} takes the form
$f_m(x)=\sum_{i=1}^Nk_m(x,x_i)\alpha_{i,m}$ ($m=1,\ldots,M$); therefore
we can equivalently solve the following finite-dimensional minimization problem:
\begin{align}
\label{eq:problem}
 \minimize_{\substack{
\valpha_m\in\RR^{N}\\(m=1,\ldots,M),\\b\in\RR} }\quad&L\Bigl(\sum_{m=1}^M\mK_m\valpha_m+b\bm{1}\Bigr)+C\sum_{m=1}^M\Bigl((1-\lambda)\|\valpha_m\|_{\mK_m}+\frac{\lambda}{2}\|\valpha_m\|_{\mK_m}^2\Bigr),
\end{align}
where $\mK_m\in\RR^{N\times N}$ is the $m$-th Gram matrix, $\valpha_m=(\alpha_{1,m},\ldots,\alpha_{N,m})\T$ is the
weight vector for the $m$-th kernel, and
$\bm{1}\in\RR^{N}$ is a vector of all one; in addition,
$L(\vz)=\sum_{i=1}^n\ell(z_i,y_i)$ . 
Moreover, we define $\|\valpha_m\|_{\mK_m}=\sqrt{\valpha_m\T\mK_m\valpha_m}$.  

The minimization problem \eqref{eq:functional-problem} is connected to the
commonly used ``{\em learning the kernel-weights}'' formulation of MKL
in the following way. First let us define
$g(x)=(1-\lambda)\sqrt{x}+\frac{\lambda}{2}x$ for $x\geq 0$ and
$g(x)=-\infty$ for $x<0$. Since $g$ is a concave function, it can be
linearly upper-bounded as $g(x)\leq xy-g^\ast(y)$, where $g^\ast(y)$ is
the concave conjugate of $g(x)$. Thus substituting
$x=\|\valpha_m\|_{\mK_m}^2$ and $y=\frac{1}{2\beta_m}$ for $m=1,\ldots,M$
in \Eqref{eq:problem}, we have:
\begin{align*}
\minimize_{\valpha_m,b,\beta_m}\quad
 &L\Bigl(\sum_{m=1}^M\mK_m\valpha_m+b\bm{1}\Bigr)+C\sum_{m=1}^M\Bigl(\frac{\|\valpha_m\|_{\mK_m}^2}{2\beta_m}-g^\ast\Bigl(\frac{1}{2\beta_m}\Bigr)\Bigr),
\intertext{where} 
g^\ast\Bigl(\frac{1}{2\beta_m}\Bigr)&=-\frac{1}{2}\frac{(1-\lambda)^2\beta_m}{1-\lambda\beta_m}.
\end{align*}
Minimizing the above expression wrt $\valpha_m$
{\em while keeping the loss term unchanged} (i.e.,
$\sum_{m=1}^M\mK_m\valpha_m=\vz$ for some $\vz$), we have
$\valpha_m=\beta_m\valpha^\ast$ and finally we can rewrite
\Eqref{eq:problem} as follows:
\begin{align*}
 \minimize_{\valpha^\ast\in\RR^n,b\in\RR,\vbeta\in\RR^M}&\quad
L\Bigl(\mK(\vbeta)\valpha^\ast+b\bm{1}\Bigr)+\frac{C}{2}\Bigl({\valpha^\ast}\T\mK(\vbeta)\valpha^\ast+\sum_{m=1}^M\tilde{g}(\beta_m)\Bigr),
\end{align*}
where $\mK(\vbeta)=\sum_{m=1}^M\beta_m\mK_m$ and
$\tilde{g}(\beta_m)=-2g^\ast(1/(2\beta_m))$. 
Therefore
\Eqref{eq:problem} is equivalent to learning the decision function with
a combined kernel $\mK(\vbeta)$ with the Tikhonov regularization on the
kernel weights $\beta_m$. Note that $\tilde{g}(\beta)=\beta$ ($\ell_1$-MKL) if
$\lambda=0$ and $\tilde{g}(\beta)$ approaches the indicator function of the
closed interval $[0,1]$ in the limit $\lambda\rightarrow 1$
(uniformly-weighted MKL). In this paper we call 
$\vbeta=(\beta_m)_{m=1}^M$ a {\em kernel-weight spectrum}.

The regularization in \Eqref{eq:functional-problem} is known as the elastic-net
regularization~\cite{ZouHas05}. In the context of 
MKL, Shawe-Taylor~\cite{Sha08} proposed a similar approach that uses the square
of the linear sum of norms in \Eqref{eq:problem}. Both Shawe-Taylor's
and our approach use mixed ($\ell_1$- and $\ell_2$-) regularization on
the {\em weight vector} (or its non-parametric version)  in the hope of curing
the over-sparseness of $\ell_1$-MKL.

 There are alternative approaches that apply non-$\ell_1$-regularization
 on the {\em kernel weights} $\beta_m$. Longworth and
Gales~\cite{LonGal09} used a combination of $\ell_1$-norm constraint and
$\ell_2$-norm penalization on the kernel weights.
Kloft \etal~\cite{KloBreSonLasMulZie10} proposed to regularize
the $\ell_p$-norm of  the kernel weights (see also \cite{CorMohRos09}). 
 Our approach (and \cite{LonGal09}) differ from
\cite{KloBreSonLasMulZie10} in that we can obtain different levels of
{\em sparsity} for all $\lambda<1$ (see bottom row of \Figref{fig:caltech}), whereas for all $p>1$ the resulting kernel-weight spectrum
is dense in \cite{KloBreSonLasMulZie10}. Note also
that uniformly-weighted MKL  ($\varphi=\infty$ in \cite{LonGal09} and $p=\infty$
 in \cite{KloBreSonLasMulZie10}) corresponds to $\lambda=1$ in our
 approach, which may be a possible advantage of our approach. 

\section{Results}
\subsection{Real data}
We computed 1,760 kernel functions on 10 binary image classification problems
(between every combinations of ``anchor'', ``ant'', ``cannon'',
``chair'', and ``cup'') from Caltech 101 dataset~\cite{FeiFerPer04}. 
The kernel functions were constructed as combinations of the following
four factors in the prepossessing pipeline:
\begin{itemize}
 \item Four types of SIFT features, namely hsvsift (adaptive scale),
       sift (adaptive scale), sift (scale fixed to 4px), sift (scale
       fixed to 8px). We used the implementation by van de Sande \etal.~\cite{vdSGevSno10}. The local features were sampled uniformly
       (grid) from each input image. We randomly choosed 200 local
       features and assigned visual words to every local features using
       these 200 points as cluster centers.
 \item Local histograms obtained by partitioning the image into
       rectangular cells of the same size in a hierarchical manner;
       i.e., level-0 partitioning has 1 cell (whole image) level-1
       partitioning has 4 cells and level-2 partitioning has 16
       cells. From each cell we computed a kernel function by
       measuring the similarity of the two local feature histograms
       computed in the same cell from two images. In addition, the
       spatial-pyramid kernel~\cite{GraDar07,LazSchPon06}, which
       combines these kernels by exponentially decaying weights, was
       computed. In total, we used 22 kernels (=one level-0 kernel + four
       level-1 kernels + 16 level-2 kernels + one spatial-pyramid
       kernel). See also \cite{GehNow09} for a similar approach. 
\item  Two kernel functions (similarity measures). We used the Gaussian
       kernel:
\begin{align*}
 k(q(x),q(x'))&=\exp\Bigl(-\sum_{j=1}^n\frac{(q_j(x)-q_j(x'))^2}{2\gamma^2}\Bigr),
\intertext{for 10 band-width parameters ($\gamma$'s) linearly spaced between
 $0.1$ and $5$ and the $\chi^2$-kernel:}
 k(q(x),q(x'))&=\exp\Bigl(-\gamma^2\sum_{j=1}^n\frac{(q_j(x)-q_j(x'))^2}{(q_j(x)+q_j(x'))}\Bigr)
\end{align*}
for 10 band-width parameters ($\gamma$'s) linearly spaced between $0.1$ and $10$,
       where $q(x),q(x')\in\mathbb{N}_+^n$ are the 
       histograms computed in some region of two images $x$ and $x'$.
\end{itemize}
The combination of 4 sift features, 22 spacial regions, 2
kernel functions, and 10 parameters resulted in  1,760 kernel functions
in total.

\begin{figure}
 \begin{center}
  \includegraphics[width=\textwidth]{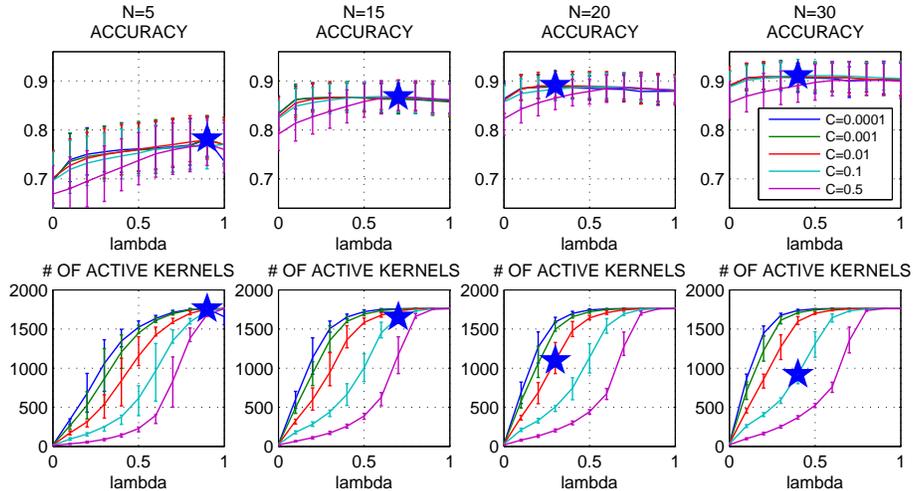}
  \caption{Image classification results from Caltech 101 dataset. The
  trade-off parameters $\lambda$ that achieve the highest test accuracy
  are marked by stars.}
  \label{fig:caltech}
 \end{center}
\end{figure}

Figure~\ref{fig:caltech} shows the average classification accuracy and the number
of active kernels obtained at different values of the trade-off parameter
$\lambda$. We can see that sparse MKL ($\lambda=0$) can be significantly
outperformed by simple uniformly-weight MKL ($\lambda=1$) when the number
of samples ($N$) is small. As the number of samples grows the difference
between the two cases decreases. Moreover, the best accuracy is
obtained at more and more sparse solutions as the number of samples
grows larger.

\subsection{Simulated data}
In order to explain the results from the image-classification dataset in
a simple setting, we generated three toy problems. In the first problem
we placed one Gaussian kernel over each input variable that was
independently sampled from the standard normal distribution. The number
of input variables was 100. We call this setting {\em Feature
selection}. In the second problem we increased the
variety of kernels by introducing 12 kernels with different band-widths on 
each input variable. The number of input variables was 10. We call this
setting {\em Feature \& Parameter selection}. In the third
problem, we used the same 12 kernel functions with different band-widths
but {\em jointly} over the same set of 10 input variables. We call this
setting {\em Parameter selection}. The true kernel-weight
spectrum $(\beta_m)_{m=1}^M$ was changed from sparse (only two non-zero
$\beta_m$'s), medium-dense (exponentially decaying spectrum) to dense
(uniform spectrum).

\begin{figure}
 \begin{center}
  \includegraphics[width=.9\textwidth]{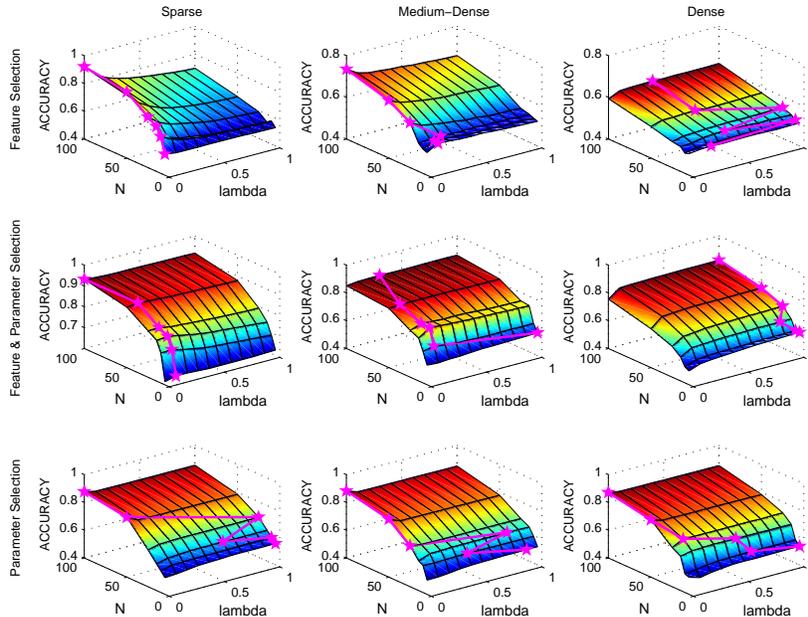}
  \caption{Classification accuracy obtained from the simulated datasets. The
  magenta colored curves with stars denote the value of trade-off
  parameters $\lambda$ that yield the highest test accuracy.}
  \label{fig:sim}
 \end{center}
\end{figure}

Figure~\ref{fig:sim} shows the test classification accuracy obtained from
training the proposed elastic-net MKL model to nine toy-problems with
different goals and different true kernel-weight spectra. We choose the
best regularization constant $C$ for each plot.
First we can observe that when the goal is to choose a subset of kernels
from {\em independent} data-sources (top row), the best trade-off parameter
$\lambda$ is mostly determined by the true kernel-weight spectrum; i.e.,
small $\lambda$ for sparse and large $\lambda$ for dense spectrum.
 Remarkably the sparse MKL ($\lambda=0$) performs well even when
the number of samples is smaller than that of kernels if the
true kernel-weight spectrum is sparse.
 On the other hand, if we also consider the selection of kernel parameter
through MKL (middle row), the best trade-off parameter $\lambda$ is
often obtained in between zero and one and seems to depend less on the
true kernel-weight spectrum. 
This finding seems to be consistent with the observation in \cite{ZouHas05} that
the elastic-net ($0<\lambda<1$) performs well when the input variables
 are linearly dependent because kernels that only differ
in the band-width can have significant dependency to each other.
Furthermore, if we consider the
selection of kernel parameter only (bottom row), the accuracy becomes
almost flat for all $\lambda$ regardless of the true kernel-weight spectrum.
 The behaviour in the Caltech
dataset seems to be most similar to the second column of the second row
(feature \& parameter selection under medium sparsity).

\section{Summary}
In this paper, we have empirically investigated the trade-off between
sparse and uniformly-weighted MKL using the elastic-net type regularization
term for MKL. The sparsity of the solution is modulated by changing
the trade-off parameter $\lambda$. We consistently found that, (a)
often the uniformly-weighted MKL ($\lambda=1$) outperforms sparse MKL
($\lambda=0$); (b) the difference between the two cases decreases as the
number of samples increases; (c) when the input kernels are independent,
the sparse MKL seems to be favorable if the true kernel-weight spectrum
is not too dense; (d) when the input kernels are linearly dependent (e.g.,
kernels with neighboring parameter values are included), intermediate
$\lambda$ value seems to be favorable. We have also observed that as the
number of samples increases the sparser solution (small $\lambda$) is
preferred. It was also observed (results not shown) that sparser
solution is preferred when the noise in the training labels is small.


\end{document}